\documentclass{article}
\usepackage{spconf,amsmath,graphicx}

\usepackage{enumitem}
\setlist{nosep, leftmargin=14pt}

\usepackage{mwe}

\usepackage{mwe} % to get dummy images
\usepackage{cite}
\usepackage{amsmath,amssymb,amsfonts}
\usepackage{algorithmic}
\usepackage{graphicx}
\usepackage{textcomp}
\usepackage{lipsum}
\usepackage{float}
\usepackage{fontenc}
\usepackage{setspace}
\usepackage{multirow}
\usepackage{textcomp}
\usepackage{bm}
\usepackage{listings}
\usepackage[mathscr]{euscript}
\usepackage{multirow}
\usepackage{float}
\usepackage[table,xcdraw]{xcolor}
\usepackage{caption}
\usepackage{setspace}
\usepackage{booktabs}
\usepackage{hyperref}

\usepackage{bibspacing}

    \title{TMI-CLNet: Triple-Modal Interaction Network for Chronic Liver Disease Prognosis from Imaging, Clinical, and Radiomic Data Fusion}
\name{
\begin{tabular}[t]{c}
Linglong Wu$^{1,\dagger}$,
Xuhao Shan$^{1,\dagger}$,
Ruiquan Ge$^{1,*}$,
Ruoyu Liang$^2$,
Chi Zhang$^2$,\\
Yonghong Li$^3$,
Ahmed Elazab$^4$,
Huoling Luo$^{3,7}$,
Yunbi Liu$^5$,
Changmiao Wang$^{6,*}$
\thanks{† These authors contributed equally to this work.\\
$\!\!\!^*$Correspondings: gespring@hdu.edu.cn, cmwangalbert@gmail.com.}
\end{tabular}}
\address{\fontsize{11pt}{5pt}\selectfont$^1$Hangzhou Dianzi University, China \quad
\fontsize{11pt}{5pt}\selectfont$^2$The Chinese University of Hong Kong, Shenzhen, Shenzhen, China \\
\fontsize{11pt}{5pt}\selectfont$^3$Shenzhen Institute of Information Technology, China$\,$
\fontsize{11pt}{5pt}\selectfont$^4$Shenzhen University, China\\
\fontsize{11pt}{5pt}\selectfont$^5$Nanjing University of Posts and Telecommunications , China
\fontsize{11pt}{5pt}\selectfont$^6$Shenzhen Research Institute of Big Data, China\\
\fontsize{11pt}{5pt}\selectfont$^7$Shenzhen Institute of Advanced Technology, Chinese Academy of Sciences, Shenzhen, China
}
\begin{document}
% \ninept
%
\maketitle
\begin{abstract}
Chronic liver disease represents a significant health challenge worldwide and accurate prognostic evaluations are essential for personalized treatment plans. Recent evidence suggests that integrating multimodal data, such as computed tomography imaging, radiomic features, and clinical information, can provide more comprehensive prognostic information. However, modalities have an inherent heterogeneity, and incorporating additional modalities may exacerbate the challenges of heterogeneous data fusion. Moreover, existing multimodal fusion methods often struggle to adapt to richer medical modalities, making it difficult to capture inter-modal relationships. To overcome these limitations, We present the Triple-Modal Interaction Chronic Liver Network (TMI-CLNet). Specifically, we develop an Intra-Modality Aggregation module and a Triple-Modal Cross-Attention Fusion module, which are designed to eliminate intra-modality redundancy and extract cross-modal information, respectively. Furthermore, we design a Triple-Modal Feature Fusion loss function to align feature representations across modalities. Extensive experiments on the liver prognosis dataset demonstrate that our approach significantly outperforms existing state-of-the-art unimodal models and other multi-modal techniques.
Our code is available at \href{https://github.com/Mysterwll/liver.git}{https://github.com/Mysterwll/liver.git}.

\end{abstract}

%

%
%%%%%%%%% BODY TEXT
\section{Introduction}
\label{sec:intro}

Chronic liver disease poses a significant threat to human health and safety. Chronic hepatitis caused by viral infections of hepatitis B virus (HBV) and hepatitis C virus (HCV) may progress to cirrhosis and hepatocellular carcinoma (HCC)\cite{campos2024update}.
Therefore, providing clinicians with timely and accurate prognostic results to guide early interventions and treatment decisions is of critical importance in mitigating the health burden of chronic liver diseases.
% Although machine learning and artificial intelligence technologies have been developed to support predictive assessments\cite{Singal2021DiagnosisAT, Zhang2022DeepLW}, they often fail to fully utilize the multidimensional patient data available in clinical settings and frequently overlook the potential relationships between different types of data. 
% These models frequently overlook the potential relationships between different types of data that could enhance overall model performance. This deficiency highlights the need for more comprehensive approaches that integrate various data modalities to improve prognostic accuracy \cite{Decharatanachart2020ApplicationOA, Musunuri2021AcuteonChronicLF}.
% In clinical practice, patients present various non-imaging data alongside routine imaging examinations such as computed tomography (CT). This non-imaging data includes basic demographic information, complete blood count reports, and lipid analysis reports \cite{Sheng2021TheUO}. 
In clinical practice, patients present various non-imaging data including demographic information, complete blood count reports, and lipid analysis reports alongside routine imaging examinations such as Computed Tomography (CT).  
Meanwhile, radiomics can extract quantitative features from medical imaging data to describe disease characteristics. By integrating comprehensive domain knowledge with auxiliary information, we can develop a holistic and precise prognostic model for chronic liver disease, thereby offering robust support for clinical practice.

Numerous studies have explored multimodal approaches across various fields \cite{yu2025prnet}
% \cite{Xue2021MultiModalCF, yu2025prnet}
, often merging different modalities through simple concatenation or attention mechanisms
\cite{khan2023multi, 10635317}.
% \cite{Musunuri2021AcuteonChronicLF, khan2023multi}.
However, these conventional methods fail to accurately capture the intricate relationships between different modalities and struggle when dealing with a larger number of data types, thereby falling short of the demands of more complex datasets. 

% In the fusion of multimodal data for prognosis assessment, two primary challenges arise. The first is the redundancy that exists within a single modality, and the second is the inherent heterogeneity between different modalities. 

To address the challenges, we pioneeringly introduce a multimodal approach for prognosis assessment in chronic liver disease. Our work makes three notable contributions: (1) To our knowledge, this is the pioneering study to simultaneously integrate CT imaging, radiomic features, and clinical information into a unified multi-modal learning framework. 
% By utilizing interaction mechanisms across these modalities, our model effectively combines complementary information to achieve more accurate prognostic assessments. 
(2) We design a unique TCAF module to address the heterogeneity among different modalities. This module effectively extracts cross-modal information to generate comprehensive global feature representations. (3) We introduce a TMFF loss to align feature representations across these three modalities during training, ensuring consistent semantic matching.
% We introduce a TMFF loss function to efficiently integrate clinical text, radiomics, and CT image features. This function aligns feature representations across these three modalities during training, ensuring consistent semantic matching.
% \begin{itemize}

% \item To our knowledge, this is the pioneering study to simultaneously integrate CT imaging, radiomic features, and clinical information into a unified multi-modal learning framework. By utilizing interaction mechanisms across these modalities, our model effectively combines complementary information to achieve more accurate prognostic assessments.

% \item We design a unique TCAF module to address the heterogeneity among different modalities. This module effectively extracts cross-modal information to generate comprehensive global feature representations.

% \item We introduce a TMFF loss function to efficiently integrate clinical text, radiomics, and CT image features. This function aligns feature representations across these three modalities during training, ensuring consistent semantic matching.

% \end{itemize}
\begin{figure*}[ht]
    \begin{center}
    \includegraphics[width=0.9\linewidth]{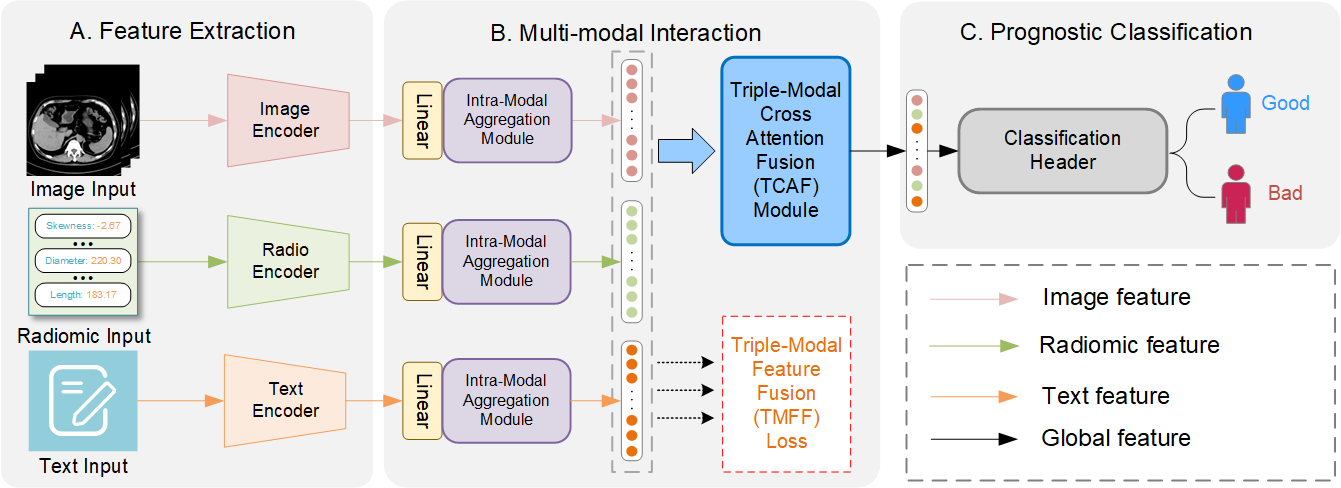}
    \end{center}
   \caption{Overview of our proposed TMI-CLNet. Given the CT scans, radiomic features, and clinical text, we feed them into an independent modal encoder to extract high-dimensional features. Subsequently, these features are fused through multi-modal interaction to incorporate global characteristics for prognosis classification. Simultaneously, the TMFF loss is employed to align multi-modal features. }
    \label{overview}
\end{figure*}
%-------------------------------------------------------------------------
\vspace{-0.15cm}
\section{Methodology}

The overall architecture of our network, as illustrated in Figure \ref{overview}, consists of three main components: a feature extraction module, a multi-modal interaction module, and a classification head module. 

In the feature extraction module, we employed a pre-trained 3D ResNet-50 as the visual encoder to extract deep visual features \cite{hara2018can}. The radiomics data were tensorized and then processed using a Multilayer Perceptron (MLP) as the radiomics encoder to achieve more abstract feature representations. For clinical text, we utilized a pre-trained BioBERT model as the text encoder, selecting the output of the last hidden layer as the text feature representation\cite{Lee2019BioBERTAP}.
In the classification head module, a pre-trained 1D DenseNet-121 is employed as the classification head.

% In the multi-modal interaction module, we employed linear layers from the outset to map the high-dimensional features obtained from the feature extraction module to a uniform size. 
% Following this, the normalized high-dimensional features are processed by the intra-modal aggregation (IMA) module to enhance their quality. At the core of the IMA is a multi-head self-attention mechanism with 16 attention heads. The primary role of the IMA is to consolidate intra-modal information before feature fusion, thereby improving the effectiveness of the fusion process and enhancing the overall performance of the model.

\subsection{Multi-modal Interaction Module}
In this module, we employed linear layers from the outset to map the high-dimensional features obtained from the feature extraction module to a uniform size. 
Following this, the normalized high-dimensional features are processed by the Intra-Modal Aggregation (IMA) module to enhance their quality. At the core of the IMA is a multi-head self-attention mechanism with 16 attention heads. The primary role of the IMA is to consolidate intra-modal information before feature fusion, thereby improving the effectiveness of the fusion process and enhancing the overall performance of the model.

\begin{figure}[h]
    \begin{center}
    \includegraphics[width=\linewidth]{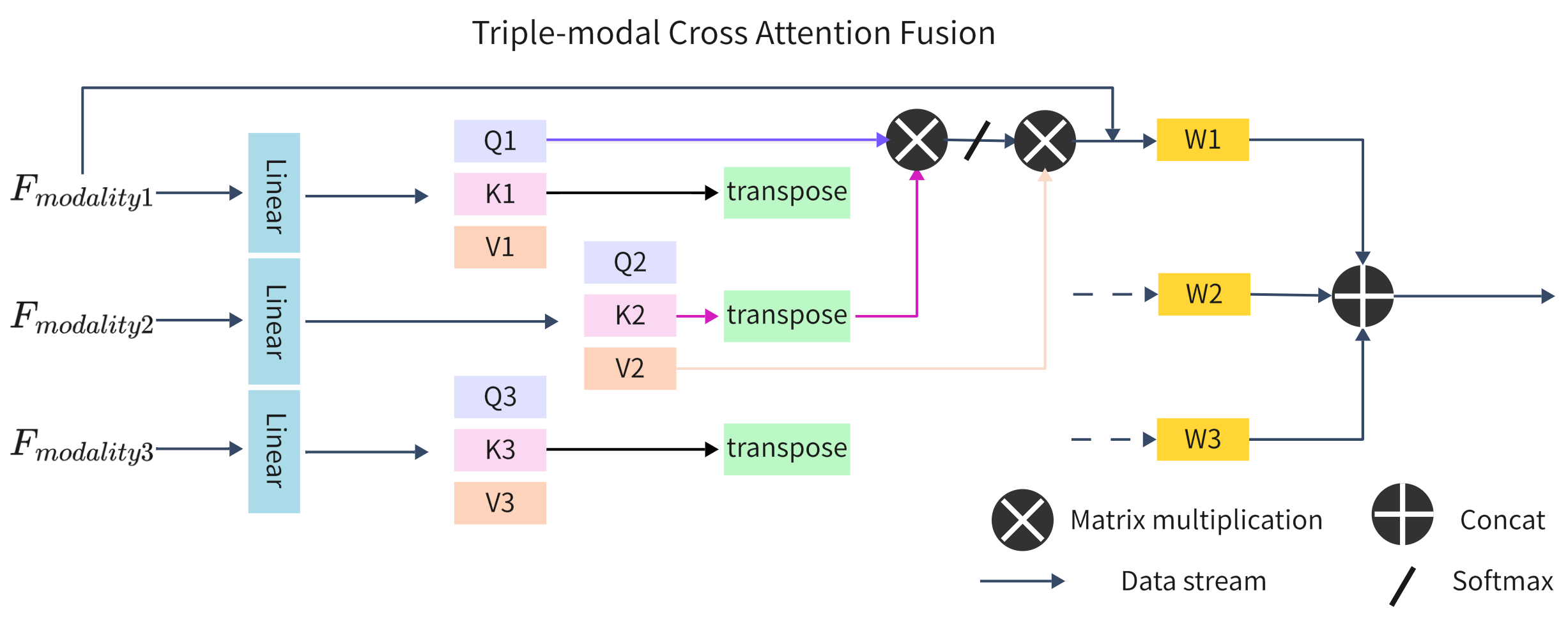}
    \end{center}
   \caption{Overview of our proposed TCAF module. The highlighting the process by the independent features from Modality 1 are integrated to produce cross-modal features.}
    \label{TCAF}
\end{figure}

\subsection{Triple-modal Cross-Attention Fusion Module}

The architecture of this module is illustrated in Figure \ref{TCAF}. After processing through the IMA module, the enhanced features from different modalities pass through linear layers to calculate their respective query \(\mathit{Q}\), key \(\mathit{K}\), and value \(\mathit{V}\)  components for attention computation. The \(\mathit{Q}\) of each modality is multiplied by the transposed \(\mathit{K}\) of the adjacent modality, followed by a softmax layer. This result is then multiplied by the \(\mathit{V}\) of the adjacent modality to compute the attention scores for the given modality. Formally, this can be expressed as:

% \begin{equation}
%     [Q_{i}, K_{i}, V_{i}] = F_{modality}^{i}W_{i},
%     \label{3}
% \end{equation}

% \begin{equation}
%     F_{hidden}^{i} = softmax(Q_{i} \times K_{j}^{T}) \times V_{j},
%     \label{4}
% \end{equation}

\begin{equation}
    F_{hidden}^{i} = softmax(\frac{Q_{i}K_{j}}{\sqrt{d_{k} } }) V_{j}, j=i \bmod 3 +1,
    \label{4}
\end{equation}

where \(\mathit{i}\) and \(\mathit{j}\) represent the index of the modality, \(\mathit{d_{k}}\) is the dimensions of the vector \(\mathit{K}\), 
% \(\mathit{F}_{modality}^{i}\) represents the modality features obtained from the different modalities after passing through the encoder and \(\mathit{W}\) represents a large weight matrix. 
Moreover, \(\mathit{a\bmod n}\) represents the remainder of \(\mathit{a}\) divided by \(\mathit{n}\). This is because the third modality interacts with the first modality through computational operations. Subsequently, to retain the original independent feature information, the attention scores (\( \mathit{F_{hidden}} \)) are combined with the initial features (\( \mathit{F_{modality}} \)) through concatenation. This combined output is then multiplied by a learnable weight matrix (\( \mathit{W^{o}} \)). 

This procedure is applied consistently to inputs from the three different modalities. Ultimately, the results are concatenated to produce the global feature output (\( \mathit{F_{global}} \)). Mathematically, this can be represented as:

\begin{equation}
    F_{global} = \bigoplus_{i=1}^{n} W_{i}^{o}  (F_{hidden}^{i} \oplus F_{modality}^{i}).
    \label{5}
\end{equation}

% \begin{equation}
%     F_{global} = Concat\left(W_{i}^{o} f_{i}^{mix}\right), i\in (1,3).
%     \label{5}
% \end{equation}

This ensures that the connected features effectively serve as the comprehensive global feature representation.

% \subsection{Classification Head Module}

% In the classification head module, a pre-trained 1D DenseNet-121 is employed as the classification head. Preliminary experiments indicate that the 1D DenseNet-121 excels at learning the features fused by the TCAF module, outperforming fully connected layers. Furthermore, leveraging pre-trained parameters helps accelerate model convergence and reduce overfitting.

\subsection{Loss Function}
\textbf{TMFF Loss. } 
To align features across different modalities, we drew inspiration from the Similarity Distribution Matching (SDM) loss \cite{jiang2023cross} and developed a TMFF loss specifically for clinical texts, radiomics, and deep image features. The SDM loss was originally designed for the global alignment of visual and textual embeddings and is defined as:

\begin{equation}
     \mathscr{L}_{\mathrm{v2t}} = KL\left(\mathbf{p}_{\mathbf{i}} \| \mathbf{q}_{\mathbf{i}}\right) = \frac{1}{n} \sum_{i=1}^{n} \sum_{j=1}^{n} p_{i, j} \log \left(\frac{p_{i, j}}{q_{i, j}}\right), 
\end{equation}

where \(q\) denotes the genuine matching probability, and \(p\) represents the fraction of a particular cosine similarity score relative to the total sum. 
The bidirectional SDM loss aggregates both \(v2t\) and \(t2v\) losses. Thus, the SDM loss between visual and textual modalities can be described as:

\begin{equation}
    \mathscr{L}_{SDM}^{vt} = \mathscr{L}_{v2t} + \mathscr{L}_{t2v}.
\end{equation}

To align the features of the three modalities, we computed their pairwise SDM losses: \(\mathscr{L}_{SDM}^{it}\) (images and text), \(\mathscr{L}_{SDM}^{rt}\) (radiomics and text), and \(\mathscr{L}_{SDM}^{ir}\) (images and radiomics). Given the asymmetry of the SDM loss across different modality pairs and the similarity between image and radiomics features, aligning images with text should correspond to aligning radiomics with text. Therefore, we define the TMFF loss as follows:

\begin{equation}
    \mathscr{L}_{multi} = \lambda \left(\frac{\mathscr{L}_{SDM}^{it} + \mathscr{L}_{SDM}^{rt}}{2}\right) + (1 - \lambda)\mathscr{L}_{SDM}^{ir},
\end{equation}

where \(\lambda \in [0, 1]\) is a scalar weight coefficient that controls the relative importance of the alignment terms.

\textbf{Overall Loss Function. } 
We employ a joint loss function to optimize the entire TMI-CLNet network. The overall loss function is formulated as a weighted sum of the task-specific and the multi-modal alignment losses. Mathematically, this can be expressed as:

\begin{equation}
    \mathscr{L}_{total} = \mathscr{L}_{task} + \alpha \mathscr{L}_{multi},
\end{equation}

where \(\mathscr{L}_{task}\) is the commonly used cross-entropy loss function, and \(\alpha\) is a weighting hyperparameter that balances the contributions of \(\mathscr{L}_{task}\) and \(\mathscr{L}_{multi}\). The value of \(\alpha\) is determined experimentally, with a typical value set to 1.

\section{Experiments}
\subsection{Experimental Setup}
\label{metric}
\noindent\textbf{Dataset. }Our study utilized a private liver prognosis dataset provided by a partner hospital, comprising 184 patients with 109 indicating good and 75 bad prognoses. For each patient, we obtained corresponding CT images, quantitative radiomics data, and diverse clinical information. 
% The CT images were uniformly resampled to dimensions of 64×512×512. 
The CT images were first preprocessed in a common pipeline of window level selection and normalization, followed by resampling to dimensions of 64×512×512.
The radiomics data extracted from the liver region encompassed 1,781 dimensions, covering aspects such as morphology, texture, and functionality. And all radiomic features were standardized to a unified range of -1 to 1. The clinical information included basic details such as gender, height, and Body Mass Index (BMI), a 23-dimensional Complete Blood Count (CBC) and biochemical index report, and a 16-dimensional fat analysis report.

% Prognosis labels for patients were determined based on adverse liver events identified through gold-standard examinations, such as liver biopsy. These adverse events included outcomes like death, bleeding, infection, and pleural or abdominal effusion. All data were meticulously annotated by professional doctors.

% \textbf{Metrics. } For our prognosis evaluation task, we employed several metrics to assess the effectiveness and robustness of the network model. These included Accuracy (ACC), which reflects the model's overall classification ability across all categories, and the Area Under the Curve (AUC), which measures the model's overall performance. Additionally, we used the F1 Score, which combines precision and recall to provide a balanced assessment of the model's accuracy and completeness.

\noindent\textbf{Implementation Details.} All experiments employed 5-fold cross-validation to validate the model's stability and generalization capabilities. The batch size was set to 2, the learning rate to 0.0001, and the number of epochs to 300. 
% To prevent overfitting, early stopping with a patience of 50 epochs was employed. 
% The Adam optimizer was used for parameter optimization. All experiments were conducted on the same platform, an Ubuntu 20.04 server equipped with an NVIDIA A100 GPU, CUDA version 11.8, and PyTorch version 2.0.
The Adam optimizer was used for parameter optimization. All experiments were conducted using an NVIDIA A100 GPU.

\begin{table}[t] 
\centering
\caption{Comparison with other methods using different modality configurations.}
\vspace{1mm}
\setlength{\tabcolsep}{0.9mm}{
\resizebox{1.0\linewidth}{!}{ 
\begin{tabular}{ccccccc}
\hline
\multicolumn{1}{c}{Method}  & \multicolumn{1}{c}{Modality}  & \multicolumn{1}{c}{ACC (\%)} & \multicolumn{1}{c}{Precision (\%)}
&\multicolumn{1}{c}{Recall (\%)} &\multicolumn{1}{c}{F1 Score}& \multicolumn{1}{c}{AUC}\\ 
\hline
Resnet-50\cite{he2016deep} & Image & 76.67 & 80.69 & 58.36 & 0.6718 & 0.7354 \\ 
ViT\cite{dosovitskiy2020image}	&Image	&63.33	&58.81	&29.75	&0.3270	&0.5933 \\ 
SVM\cite{Mukherjee2022RadiomicsBasedML}	&Radiomic	&57.01	&47.72	&39.34	&0.4289	&0.5718\\
RF\cite{Breiman2001RandomF}	&Radiomic	&66.22	&61.82	&49.98	&0.5485	&0.6828\\ \hline

SimpleFF\cite{choi2021fully} & I+R &	72.75	&74.72&	54.10&	0.6218&	0.7050\\ 
HFBSurv\cite{li2022hfbsurv}	 &I+R+C	&79.45	&82.01	&64.32	&0.7233	&0.8015\\
MMD\cite{cui2022survival}	& I+R+C	&77.22	&81.25	&60.63	&0.6772	& 0.7344 \\ \hline

\textbf{TMI-CLNet (Our)} & I+R+C &\textbf{83.12}	&\textbf{84.38}	&\textbf{74.28}	&\textbf{0.7805}	&\textbf{0.8223} \\ \hline

\end{tabular}}}
\label{comparison}
\vspace{-1.5em}
\end{table}

\subsection{Experimental Results}
\textbf{Comparison with Other Methods.} 
Table \ref{comparison} presents the comparative results of various methods. The first four methods rely on single-modal data as baselines, employing classic deep learning techniques such as ResNet50\cite{he2016deep} and ViT\cite{dosovitskiy2020image} for images, and machine learning techniques like Support Vector Machine (SVM)\cite{Mukherjee2022RadiomicsBasedML} and Random Forest (RF)\cite{Breiman2001RandomF} for radiomics data. The next three methods represent advanced approaches for handling multi-modal data. SimpleFF\cite{choi2021fully} sequentially integrates deep learning and radiomics data. HFBSurv\cite{li2022hfbsurv} progressively fuses multi-modal data using a factorized bilinear model. 
Lastly, MMD\cite{cui2022survival} offers a universal framework for multi-modal feature fusion, accommodating both complete and missing modalities.

% For the general medical multi-modal feature fusion frameworks HFBSurv and MMD, we retained our feature extraction encoders to adapt to the different data formats inherent in various modalities. 
% Among the single-modal data approaches, ResNet50 significantly outperforms other methods, demonstrating its robustness and effectiveness in medical imaging. Traditional machine learning methods generally yield lower performance, likely due to redundancy within radiomics data.
Integrating CT images with other modalities can sometimes degrade performance such as SimpleFF. 
% For example, SimpleFF achieves an accuracy of 72.75\%, an F1 Score of 0.6218, and an AUC of 0.7050, which are lower by 3.92\%, 0.0500, and 0.0304, respectively, compared to ResNet50 on images alone. 
This decline may be attributed to the inherent heterogeneity among different modalities, posing challenges for effective multi-modal feature fusion. In contrast, HFBSurv considers modality heterogeneity and cross-modal relationships, surpassing all single-modality methods. Our approach demonstrates superior performance across all metrics compared to other methods. Specifically, our framework achieved an accuracy of 83.12\%, an F1 score of 0.7805 and an AUC of 0.8223. These results represent improvements of 3.67\%, 0.0572, and 0.0208, respectively, over the best metrics from other methods. 

% Furthermore, as illustrated in Figure \ref{auc_curve}, the ROC curve approaches the upper left corner, substantiating the superiority of our proposed multi-modal interaction framework.

\textbf{Ablation Study for Modalities.}
Each experiment included the relevant encoders and classification heads. We used a direct connection for unimodal fusion, whereas, for bimodal fusion, we employed cross-attention mechanisms.
As shown in Table \ref{Modalities}, the results from the bimodal ablation experiments were similar to or slightly lower than those obtained with CT imaging alone. 
% This might be due to redundant information in our original data that the cross-attention structure in the bimodal setup did not effectively filter out.
Specifically, the combination of vision and radiomics performed poorly, likely due to inconsistencies in the representation and dimensionality between deep features and handcrafted features, which simple cross-attention mechanisms are unable to effectively handle.
% However, our method achieved the best performance, improving accuracy by 5.44\% and AUC by 6.82\% compared with the best-performing unimodal method. This supports the effectiveness of our approach in handling modality heterogeneity and leveraging cross-modal information.
However, our method achieved the best performance and this supports the effectiveness of our approach in handling modality heterogeneity and leveraging cross-modal information.

\begin{table}[h] 
\centering
\caption{Results of modal ablation experiment.}
\vspace{1mm}
\setlength{\tabcolsep}{0.9mm}{
\resizebox{1.0\linewidth}{!}{
\begin{tabular}{cccccc}
\hline
\multicolumn{1}{c}{Method}  & \multicolumn{1}{c}{ACC (\%)} & \multicolumn{1}{c}{Precision (\%)}
&\multicolumn{1}{c}{Recall (\%)} &\multicolumn{1}{c}{F1 Score}& \multicolumn{1}{c}{AUC}\\ 
\hline
Vision-Only	&77.68	&78.53	&64.98	&0.6983	&0.7541\\
Text-Only	&73.33	&74.91	&54.21	&0.6199	&0.7001\\
Radio-Only	&72.79	&69.68	&61.91	&0.6459	&0.7123\\ 
Vision+Text	&76.64	&71.43 	&68.75 	&0.6276	&0.7224\\ 
Radio+Text	&77.16	&82.51	&58.62	&0.6707	&0.7409\\ 
Vision+Radio &74.94 &78.77  &56.95  &0.6341 &0.7218\\ 

\hline

\textbf{TMI-CLNet (Our)}  &\textbf{83.12}	&\textbf{84.38}	&\textbf{74.28}	&\textbf{0.7805}	&\textbf{0.8223} \\ \hline

\end{tabular}}}
\label{Modalities}
\vspace{-0.25em}
\end{table}
\textbf{Ablation Study for IMA and TCAF. }
To further validate the effectiveness of our modules in reducing intra-modality redundancy and extracting cross-modal information, we conducted ablation experiments on the IMA and TCAF modules. The results, presented in Table \ref{component}, show a significant performance improvement when each module was integrated individually. When both IMA and TCAF modules were used together, the accuracy increased by 7.64\%. These results demonstrate that the network effectively integrates multimodal information when all modules are applied, highlighting the strength of our approach in handling complex data.

\begin{table}[h] 
\centering
\caption{Ablation studies of important modules.}
\vspace{1mm}
\setlength{\tabcolsep}{0.8mm}{
\resizebox{1.0\linewidth}{!}{
\begin{tabular}{ccccccc}
\hline
\multicolumn{1}{c}{IMA}  & \multicolumn{1}{c}{TCAF}  &\multicolumn{1}{c}{ACC (\%)} & \multicolumn{1}{c}{Precision (\%)}
&\multicolumn{1}{c}{Recall (\%)} &\multicolumn{1}{c}{F1 Score}& \multicolumn{1}{c}{AUC}\\ 
\hline
  - &  - & 75.48	&75.12	&63.34	&0.6504 &0.6952	\\
   &  \checkmark & 78.89	&82.79	&62.92	&0.7019	&0.7686\\
\checkmark  &   &81.47	&83.94	&68.55	&0.7491	&0.7883\\
\checkmark	&  \checkmark &\textbf{83.12}	&\textbf{84.38}	&\textbf{74.28}	&\textbf{0.7805}	&\textbf{0.8223}\\ \hline

\end{tabular}}}
\label{component}
\vspace{-0.25em}
\end{table}

\textbf{Ablation Study for Loss Weights. }
In the TMFF loss, we introduce weight coefficients $\lambda$ to control the relative importance of individual pairwise SDM losses. This ablation study examines the impact of different weight coefficients. The experimental results are shown in Table \ref{loss}, where "-" indicates the absence of TMFF loss. By employing the TMFF loss and adjusting the coefficients $\lambda$ to 0.2, 0.4, 0.6, and 0.8, we observe varying degrees of performance improvements.
When $\lambda$ is set to 0.6, our model surpasses the baseline without TMFF loss by 6.45\% in accuracy and 0.0687 in AUC. These results underscore the significance of TMFF loss in aligning and integrating features across different modalities. 

% Although our method does not achieve the best precision and recall at $\lambda = 0.6$, we still opt for 0.6 as the optimal parameter for TMFF loss as the F1 score and AUC typically reflect the model’s comprehensive capabilities.

 \begin{table}[h]
\centering
\caption{Ablation studies of weight coefficients $\lambda$ in the TMFF Loss. }
\vspace{1mm}
\setlength{\tabcolsep}{0.9mm}{
\resizebox{1.0\linewidth}{!}{ 
\begin{tabular}{cccccc}
\hline
\multicolumn{1}{c}{$\lambda$}  & \multicolumn{1}{c}{ACC (\%)} & \multicolumn{1}{c}{Precision (\%)}
&\multicolumn{1}{c}{Recall (\%)} &\multicolumn{1}{c}{F1 Score}& \multicolumn{1}{c}{AUC}\\  \hline
-   &76.61  &73.42  &68.36  &0.7050 &0.7536  \\
0.2	&80.95 	&79.66	&\textbf{75.13}	&0.7636	&0.8017  \\
0.4	&78.21	&76.93	&67.33	&0.7141	&0.7668  \\
0.6	&\textbf{83.12}	&84.38	&74.28	&\textbf{0.7805}	&\textbf{0.8223}  \\
0.8	&79.85	&\textbf{87.31}	&59.78	&0.7028	&0.7659  \\ \hline

\end{tabular}}}
\label{loss}
\vspace{-0.25em}
\end{table}

\textbf{Interpretative Visualization.} In this study, we utilized XGradCAM \cite{selvaraju2017grad} to analyze the decision-making process of our proposed TMI-CLNet. We selected four intermediate layers of the visual encoder as target layers, averaged their class activation maps, and converted them into heatmaps. Compared to a purely visual encoder, our fusion network focuses more on the liver region and reduces attention to task-irrelevant tissue areas. This indicates that, guided by clinical and radiomic information, the model excels in extracting relevant features from CT data.
\begin{figure}[h]
    \centering
    \includegraphics[width=\linewidth]{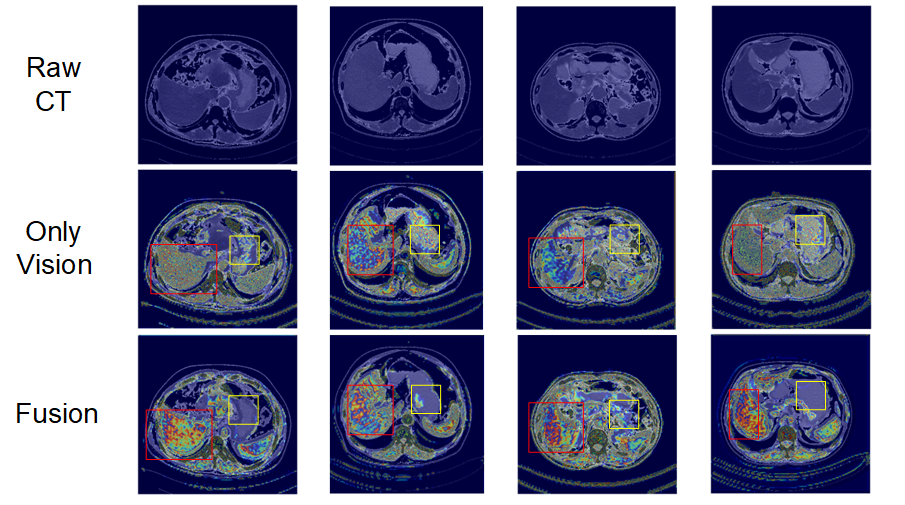}
    \caption{Visual examples illustrating the model's focus before and after fusion. The redder areas indicate higher model attention. Red boxes highlight the liver region, whereas yellow boxes denote areas deemed irrelevant by the model.}
    \label{cam}
\end{figure}

% \begin{figure}[t]
%     \centering
%     \includegraphics[width=\linewidth]{pdf/auc.png}
%     \caption{ROC curves between TMI-CLNet and other methods.}
%     \label{auc_curve}
% \end{figure}

\section{Conclusion}
This study presented TMI-CLNet, which integrates CT imaging, radiomic features, and clinical infomation to provide early prognosis assessments for patients with chronic liver disease. By introducing the TCAF module and the TMFF loss function, the proposed model can address the heterogeneity among different modalities, thus achieving remarkable performance. 
% Ablation experiments and comparative studies with other multi-modal methods demonstrate the efficacy of our method.
Experimental results demonstrate the efficacy of our method. In the future, we will focus on expanding the dataset, conducting multi-center validation, and improving computational efficiency to enhance the robustness and scalability of the model. Additionally, our approach provides a valuable perspective and can be extended to other diseases or modalities, making it suitable for various future applications across different fields.

% These findings have the potential to assist clinicians in making more accurate early-stage prognostic assessments for chronic liver disease patients, ultimately improving treatment outcomes.
%-------------------------------------------------------------------------
\section{COMPLIANCE WITH ETHICAL STANDARDS}
\vspace{-0.6em}
This study was conducted in accordance with the principles of the Declaration of Helsinki. Approval was granted by the Ethics Committee of Longgang Central Hospital of Shenzhen (2024.5.8/No.2024052).

\vspace{-0.6em}
\section{ACKNOWLEDGMENTS}
\vspace{-0.6em}
This work was supported by the Open Project Program of the State Key Laboratory of CAD\&CG (No.A2410), Zhejiang University,   Zhejiang Provincial Natural Science Foundation of China (No.LY21F020017), National Natural Science Foundation of China (No.61702146, 62076084), GuangDong Basic and Applied Basic Research Foundation (No.2022A1515110570), Innovation Teams of Youth Innovation in Science and Technology of High Education Institutions of Shandong Province (No.2021KJ088), and Guangdong Natural Science Foundation (No. 2023A1515012587).

% References should be produced using the bibtex program from suitable
% BiBTeX files (here: strings, refs, manuals). The IEEEbib.bst bibliography
% style file from IEEE produces unsorted bibliography list.
% ------------------------------------------------------------------------- 
\vspace{-0.6em}
\bibliographystyle{IEEEbib}
\bibliography{refs}

\end{document}